# Evaluating the Stability of Deep Learning Latent Feature Spaces

Ademide O. Mabadeje[a, *] and Michael J. Pyrcz[a, b]
[*]Corresponding author E-mail: ademidemabadeje@utexas.edu

[a]Hildebrand Department of Petroleum & Geosystems Engineering, The University of Texas at Austin, 200 East Dean Keeton Street, Stop C0300, Austin, Texas, 78712, USA.
[b]Department of Planetary and Earth Sciences, The University of Texas at Austin, 2305 Speedway, Stop C1160, Austin, Texas, 78712, USA.

## Abstract

High-dimensional datasets present substantial challenges in statistical modeling across various disciplines, necessitating effective dimensionality reduction methods. Deep learning approaches, notable for their capacity to distill essential features from complex data, facilitate modeling, visualization, and compression through reduced dimensionality latent feature spaces, have wide applications from bioinformatics to earth sciences. This study introduces a novel workflow to evaluate the stability of these latent spaces, ensuring consistency and reliability in subsequent analyses. Stability, defined as the invariance of latent spaces to minor data, training realizations, and parameter perturbations, is crucial yet often overlooked.

The proposed methodology delineates three stability types, sample, structural, and inferential, within latent spaces, and introduces a suite of metrics for comprehensive evaluation. We implement this workflow across 500 autoencoder realizations and three datasets, encompassing both synthetic and real-world scenarios to explain latent space dynamics. Employing k-means clustering and the modified Jonker-Volgenant algorithm for class alignment, alongside anisotropy metrics and convex hull analysis, we introduce adjusted stress and Jaccard dissimilarity as novel stability indicators.

The results highlight inherent instabilities in latent feature spaces and demonstrate the workflow's efficacy in quantifying and interpreting these instabilities. This work advances the understanding of latent feature spaces, promoting improved model interpretability and quality control for more informed decision-making for diverse analytical workflows that leverage deep learning.

# 1. Introduction

With the fourth paradigm of scientific discovery at its peak, data-driven, deep learning algorithms and associated latent feature spaces have been applied across various domains, including anomaly identification (Angiulli et al., 2023; Hernandez-Mejia et al., 2024; Norlander and Sopasakis, 2019), spatial, subsurface model generation (Grana et al. 2019; Jo et al. 2020; Pan et al. 2021; Feng et al. 2022; Misra et al. 2022), flow through porous media modeling (Karimpouli and Tahmasebi 2019; Santos et al. 2020; Maldonado-Cruz and Pyrcz 2022), data analysis and pattern detection in big data (Babaei et al. 2019; Chadebec and Allassonnière 2021; Pintelas and Pintelas 2022), facial recognition and reconstruction (Abdolahnejad and Liu 2022), medical diagnosis (Sobahi et al. 2022), inversion to build spatial, subsurface models that honor dynamic outputs (Liu and Durlofsky 2021; Razak et al. 2022), and clustering of analogous samples to improve inference in finance and economics (Gu et al. 2020; Mabadeje and Pyrcz 2022; MacKinnon et al. 2023). Latent feature space, also referred to as latent space is a low-dimensional representation, generally a vector of standardized values of limited length, which adequately describes a high-dimensional dataset to enhance model efficiency and interpretability. In many cases, there is a mapping from this latent space to a unique model in the original high-dimensional space.

High-dimensional data describes datasets in which the number of dimensions, $d$, exceeds the number of observations or samples, $N$. Each dimension represents a feature in tabular datasets or a cell in image datasets. High-dimensional data, characterized by dimensions, $d$, and data samples, $N$, in $\mathbb{R}^{N \times d}$, poses significant challenges in any statistical modeling and analyses (Sorzano et al. 2014) due to the curse of dimensionality. The curse of dimensionality refers to a series of statistical consequences for high-dimensional data characterized by sparse sampling, multicollinearity, and poor sampling coverage (Bellman 1957). These challenges affect the reliability and accuracy of models and systems, which often struggle with high dimensions, leading to challenges in machine learning such as reduced effectiveness of distance measures critical for assessing similarity or dissimilarity; therefore, dimensionality reduction is imperative (Aggarwal et al. 2001; Kabán 2012; Giannella 2021).

Dimensionality reduction, a branch of inferential machine learning, is the projection of high-dimensional datasets into lower dimensionality while attempting to minimize the loss of information, typically applied in extracting relevant feature information, data modeling, compression, and visualization (Van Der Maaten et al. 2009). With dimensionality reduction methods, there is an a priori assumption that every high-dimensional dataset exists on a lower-dimensional manifold (Goodfellow et al. 2016). Dimensionality reduction strategies are of two categories, linear and nonlinear dimensionality reduction methods, implying the high-dimensional dataset resides on either a linear or nonlinear manifold or data spaces, respectively. Examples of linear dimensionality reduction methods are locality-preserving projections (He and Niyogi 2004; He et al. 2005), Fisher's linear discriminant analysis (Fisher 1936; Rao 1948), principal component analysis (Pearson 1901; Jolliffe 2005), and variants utilizing kernel mappings or regularizations on a linear manifold (Schölkopf et al. 1997; Sarma et al. 2008; Kao and Van Roy 2013; Turchetti and Falaschetti 2019). However, such methods do not adequately capture the complex and nonlinear nature of many subsurface datasets because of complicated geological or engineering physics; therefore, there is a preference for nonlinear dimensionality reduction methods in these fields.

Nonlinear dimensionality reduction methods may be divided into two groups based on whether local or global structures within high-dimensional datasets are preserved. Prominent nonlinear dimensionality reduction methods preserving local structures include local linear embeddings (LLE) (Roweis and Saul 2000), t-distributed stochastic neighbor embedding (t-SNE) (van der Maaten and Hinton, 2008), and Laplacian Eigenmaps (Belkin and Niyogi 2003). For global structure preservation, methods such as multidimensional scaling (MDS) (Torgerson 1952; Borg and Groenen 2005; Cox and Cox 2008), and isometric mapping (IsoMap) (Tenenbaum et al. 2000) are commonly applied. Within this framework, Cunningham and Ghahramani (2015) demonstrate nonlinear local dimensionality reduction methods can be conceptualized as a broader variant of multidimensional scaling (MDS). However, Bengio and Monperrus (2005) highlighted the necessity for extensive training data in cases where the underlying manifolds are not smooth, thus pointing out the limitations in the generalization capabilities of such methods when faced with unseen manifold variations. This challenge spurred the development of deep learning-based techniques that attempt to preserve both local and global structures (Brunton et al. 2016; Turchetti and Falaschetti 2019; Fries et al. 2022; He et al. 2023). Prominent among these are autoencoders and their variants, such as variational autoencoders (Kingma and Welling 2019) and conditional autoencoders (Sohn et al. 2015).

Leveraging the concept that data often resides near a lower dimensional manifold or a subset of manifolds, autoencoders are designed to discern the underlying manifold structure. An autoencoder is a neural network employed in unsupervised learning, functioning by encoding high-dimensional data into a compressed latent space and subsequently reconstructing the original input from this representation (Rumelhart et al. 1986; Kramer 1992; Bengio and Lecun 2007; Hinton 2007). The architecture comprises two main components: an encoder, which maps input data, $X$, to a latent representation, $Z$, through $Z = f(X)$, and a decoder that attempts to reconstruct the input from the latent space as $X' = g(Z)$. The utility of an autoencoder lies not in the reconstruction itself, but in the latent feature space, $Z$, that ideally captures the most salient features of the data. The training of an autoencoder involves minimizing a loss function $L(X, X')$ such as mean squared error (continuous feature) or binary cross entropy (binary, categorical feature), to minimize the reconstruction error of the output, $X'$, relative to the original input, $X$. However, to prevent trivial identity mapping and ensure meaningful feature learning, autoencoders are often designed to be undercomplete, with the latent feature space's dimensionality less than the input or regularized to impose constraints on the encoding process. Regularized autoencoders, such as variational autoencoders (Kingma and Welling 2019) or generative stochastic networks (Goodfellow et al. 2014), introduce additional regularization terms or modify the architecture to encourage the model to learn useful data representations even in overcomplete settings, where the latent space dimensionality equals or exceeds the input dimension.

Given that autoencoders are specialized feedforward networks with hidden layers, its capabilities are encompassed by the universal approximation theorem (Hornik et al. 1989, 1990; Cybenko 1989). This theorem asserts that a feedforward network with at least a singular hidden layer and a linear output layer, equipped with any activation function, can approximate any function between finite-dimensional vector spaces to a specified degree of accuracy, contingent on a sufficient number of hidden units. This suggests that for any given function, a sufficiently expansive neural network can be constructed to represent it. While the universal approximation theorem confirms the potential for a network of adequate size to attain any desired level of precision, it remains silent on the exact dimensions required for such a network. Barron (1993) offers insights into the scale a single-layer network might need to approximate a wide array of functions, suggesting that an extremely large number of hidden units may be necessary in certain scenarios (Goodfellow et al. 2016). Therefore, although a single-layer feedforward network theoretically suffices to embody any function, practical limitations such as size feasibility and learning efficacy may arise. Conversely, adopting deeper architectures, that is, neural network models with a higher number of layers, including hidden layers, could potentially diminish the requisite number of units to represent a specific function and might lower generalization errors, albeit without guarantees (Goodfellow et al. 2016).

However, the more complex deep learning architectures become when used for dimensionality reduction methods, the more the stability and reliability of the attained latent feature spaces are called into question. Mabadeje and Pyrcz (2023) introduce the concept of stability in nonlinear local dimensionality reduction methods to investigate how Euclidean transformations, i.e., translation, rotation, and reflection, alters data representation in MDS latent feature space, and its mitigation via rigid transformations. Stability is defined as the property of latent feature space to remain consistent and unchanged under small perturbations or variations in the input data, multiple models runs with the same data or algorithm parameters (Dayawansa 1992). In deep learning approaches, the latent feature spaces for some methods are either translational or rotational invariant, but no existing method is invariant to all Euclidean transformations at once (Fukushima 1980; LeCun et al. 1998; Krizhevsky et al. 2017). However, instability in deep learning latent feature spaces can arise from even the slightest changes in parameters such as weight initialization, which is based on specific random states within its architecture. Although deep learning methods are based on stochastic training optimization that may sometimes have poor convergence, there is an implicit assumption that for a given dataset and model with the same architecture regardless of whether its latent feature space is Euclidean transformation variant or not, the predictive, modeling, or inferential analysis converge to an approximate solution space. Therefore, it is imperative to incorporate diagnostics that evaluate the stability and interpretability of latent feature spaces within workflows across any discipline that relies on latent feature spaces for modeling and analysis.

We propose a novel generalizable workflow, which includes a set of metrics that summarize the latent feature space as not only a diagnostic tool, but also part of the minimum acceptance criteria to evaluate the stability of deep learning latent feature spaces for robust use within any latent feature space-based workflows. To quantitatively assess the stability of these spaces, we employ k-means clustering in the autoencoders' latent space and map the encoded representations to predefined clusters in the categorical response feature, adjusting for any discrepancies through the modified Jonker-Volgenant algorithm for linear sum assignment. Next, we introduce a geostatistics-based measure, anisotropic ratio, to compute local and global anisotropy metrics to provide a spatial view of data structures within the latent space. Then, a geometric perspective on the data's dispersion and boundary is computed via convex hulls for each latent space, and anchor sets consisting of data samples on the hull are obtained. The introduction of these geostatistical and geometric-based measures allows for an intuitive understanding of the latent space dynamics in relation to the original high-dimensional data resulting in three types of stability, sample, structural, and inferential. Lastly, to ensure consistent inferences and interpretation, we use our proposed adjusted stress metric as a measure of distortion between each autoencoder latent feature space and quantify the stability of these spaces based on Jaccard dissimilarity between all anchor sets.

The background section provides an overview of existing literature on autoencoder's latent feature space applications and its connections to geometry-based statistics, optimization, and geostatistical concepts. The methodology section outlines the computation of our proposed distortion and geometry-based metrics, which are used to investigate and highlight the stability or instability of deep learning latent feature spaces. The results and discussion sections include details about the autoencoder's architecture, the generation of synthetic datasets used for workflow calibration, and the impact of correlation strength within the training dataset on latent feature spaces. Additionally, we employ the widely used wine dataset (Aeberhard and Forina 1991) in our case study analysis of the latent feature spaces, while also discussing the underlying assumptions and associated limitations of our proposed workflow. Finally, we assess the accuracy and validity of the proposed workflow for exploring the stability and interpretability of deep learning latent feature spaces obtained from high-dimensional datasets.

## 2. Background

### 2.1. *Latent feature space applications*

For subsurface process modeling in geothermal energy, carbon storage, and hydrocarbon reservoirs, large subsurface models are essential. The computational complexity, time, and memory storage, for data assimilation, engineering optimization, and forecasting scale with model size. To address these challenges, deep learning methodologies, particularly autoencoders, are employed to derive geologically coherent low-dimensional representations of subsurface models. Misra et al. (2022) introduced a vector quantized variational autoencoder approach, facilitating efficient compression and reconstruction of subsurface model layers, thereby enhancing computational efficiency in subsurface modeling. Autoencoders are used for dimensionality reduction to extract a limited set of informative features to understand the behavior of subsurface systems, such as CNN-PCA for complex three-dimensional subsurface models (Liu and Durlofsky 2021), and CNN-based latent space inversion (Razak et al. 2022) as a deep learning inversion framework that preserves the subsurface's geological attributes while integrating flow data from data assimilation to minimize the discrepancies between observed and simulated production data based on prior models.

Data assimilation, also known as historical production matching, necessitates the modification of geological parameters like permeability and porosity to align flow simulation results with observed data. Data assimilation benefits from geological parameterization, which translates high-dimensional data into a manageable set of uncorrelated low-dimensional variables preserving essential geological structures. Xiao et al. (2024) develop a hybrid neural network proxy model merging a recurrent autoencoder (R-AE) with a fully connected neural network to enable robust and efficient optimization during data assimilation to aid reliable decision-making in the subsurface. Zhu and Zabaras (2018) develop a Bayesian proxy model for uncertainty quantification and propagation in high-dimensional problems with limited training data using a deep convolutional encoder-decoder network to improve predictive accuracy and uncertainty estimates applicable to ensemble methods, Gaussian, and non-Gaussian processes.

In the field of computer vision, convolutional autoencoders are widely used in image classification, reconstruction, augmentation problems for image and face recognitions, and medical image diagnosis, for feature extraction and predictive classification (Abdolahnejad and Liu, 2022; Pintelas and Pintelas, 2022; Sobahi et al., 2022). Karimpouli and Tahmasebi (2019) employ the use of convolutional autoencoders as a stochastic image generator to create multiple realizations of digital rock images in the presence of small sample sizes to enhance pore space and mineralogy segmentation in digital rock images. Variational autoencoders are utilized as generative models to perform data augmentation for regression analyses and image generation tasks when data size is limited (Ohno 2020; Chadebec and Allassonnière 2021). Ohno (2020) focuses on regression problems related to the calculation of continuous input-output mappings in real-valued spaces to improve the generalization performance of linear regression models with augmented training data in the field of materials informatics. To combat the effect of having limited sample sizes in classification and clustering analyses, Babaei et al. (2019) use autoencoders to derive meaningful features from high-dimensional datasets while simultaneously performing data augmentation for one-class classifiers for anomaly detection.

Anomaly detection often used interchangeably as outlier detection, is a prevalent challenge in disciplines such as cyber security, medical imaging, fraud detection, and earth sciences. Anomalies are defined either as samples that occur in regions of low probability within a dataset, or outlying points of interest or deviation from expectation. Norlander and Sopasakis (2019) created a conditional latent space variational autoencoder to improve preprocessing for clustering and anomaly detection on unlabeled data, where anomaly detection is performed within the space of reduced dimensionality i.e., latent feature space. Corizzo et al. (2019), Guo et al. (2018), and Zhang et al. (2018) propose latent space-reliant anomaly detection frameworks via autoencoders for nonlinear process monitoring and predictive modeling tasks from stream data by mapping points to their latent representation and assigning them a score based on the distances from their k-nearest neighbors in the latent space. Angiulli et al. (2023) propose the use of combined autoencoder latent and reconstruction spaces for unsupervised anomaly detection over the traditional method based on reconstruction error. Hernandez-Mejia et al. (2024) proposed the use of a probabilistic convex hull based on joint probabilities in autoencoder latent feature space for the identification of both labeled and unlabeled outliers within high-dimensional datasets.

## 2.2. Convex Hull

The quickhull algorithm is a geometry-based approach proposed by Barber et al. (1996) utilizing the divide-and-conquer technique for convex hull determination in $\mathbb{R}^d$. Divide-and-conquer refers to a problem-solving approach where a complex problem is broken down into smaller, more manageable subsets, which are then solved individually. The solutions to these subsets are combined to form the solution to the original problem. Starting with the identification of extremal points that define the initial convex boundary, the algorithm recursively partitions the set of points by determining the furthest point from the existing hull segments and iteratively discarding interior points, thus refining the hull boundary. This process is repeated until all exterior points are incorporated into the hull i.e., no points remain outside the hull, yielding a sequence of segments or planes that form the convex hull polygon. For a given set $S$ of points in space, the convex hull, $H(S)$, can be represented as Eq. (1).

$$H(S) = \bigcup_{i=1}^{n} [a_i, a_{i+1}] \tag{1}$$

where $[a_i, a_{i+1}]$ denotes the line segment between successive points $a_i$ and $a_{i+1}$ on the hull, $n$ is the total number of points defining the hull, and the convex hull, $H(S)$, consists of an anchor set consisting of all anchor points (i.e., points on the hull) denoted by the mathematical set $\{\mathbf{A}_n\}$.

### 2.3. Minimum volume enclosing ellipse

Minimum volume enclosing ellipse (MVEE) is a method that relies on linear algebra, optimization, and geometric principles to encapsulate a set of data samples within the smallest possible ellipse, capturing the spread and orientation of the dataset, which is critical in understanding the underlying structures in fields like data mining, machine learning, and pattern recognition. There exist different formulations of MVEE (Silverman and Titterington 1980; Sun and Freund 2004; Kumar and Yildirim 2005; Todd and Yildirim 2007; Rosa and Harman 2022; Bowman and Heath 2023), but the focus in this work is the Khachiyan algorithm (Khachiyan 1996). The Khachiyan algorithm for computing the MVEE begins by augmenting a set of $d$-dimensional data samples into a matrix, $\boldsymbol{Q}$, with an additional row of ones to accommodate the ellipse's offset. Next, an initial uniform weight vector, $\boldsymbol{u}$, is established, dictating the influence of each sample. The algorithm iteratively refines $\boldsymbol{u}$ by computing a weighted covariance matrix of the data sample, $\boldsymbol{\Sigma}$, as $\boldsymbol{\Sigma} = \boldsymbol{Q}\text{diag}(\boldsymbol{u})\boldsymbol{Q}^T$, representing a weighted quadratic form of the points. The key step involves updating $\boldsymbol{u}$ based on matrix $\boldsymbol{M} = \text{diag}(\boldsymbol{Q}^T\boldsymbol{\Sigma}^{-1}\boldsymbol{Q})$, specifically by increasing the weight of the sample corresponding to $\boldsymbol{M}$'s maximum value while proportionally decreasing the others, maintaining their sum as unity. This iterative process continues until the adjustment in $\boldsymbol{u}$ falls below a predetermined tolerance level, indicating convergence. The final ellipse is characterized by its center, $\boldsymbol{c}$, calculated as the weighted average of the points using the optimized $\boldsymbol{u}$, and a positive definite matrix that defines the ellipse, $\boldsymbol{B}$, derived from the weighted covariance of the points relative to $\boldsymbol{c}$. Therefore, the resultant ellipse encapsulating the data samples with minimal volume is of the form $(\boldsymbol{x} - \boldsymbol{c})^T\boldsymbol{B}(\boldsymbol{x} - \boldsymbol{c}) = 1$, where $\boldsymbol{x}$ represents a generic data sample in $\mathbb{R}^{d \times n}$.

### 2.4. Modified Jonker-Volgenant Algorithm

The linear sum assignment problem (LSAP) is the task of finding a matching between two groups or sets that minimizes the sum of pairwise assignment costs. LSAP, often encountered in optimization and operations research is solved using the auction algorithm described in Bertsekas (1988), which involves assigning an equal number of agents and tasks based on a cost matrix, $\boldsymbol{C}$, where $C_{ij}$ is the cost of assigning the $i^{th}$ agent to the $j^{th}$ task to minimize the total assignment cost. However, auction algorithms do not always yield the best solution given its computational complexity depends on both the size of the problem but also the relative values of the cost matrix's elements. This gives rise to the class of Hungarian algorithms and its variants originating from Kuhn (1955), whose computational complexity depends only on the size of the problem. For example, the Jonker-Volgenant algorithm (Jonker and Volgenant 1988), is an efficient variant that has been generalized, optimized, and widely accepted as a better alternative to the auction algorithm. Crouse (2016) presents an implementation of the Jonker-Volgenant algorithm without the need for initialization that involves setting up a cost matrix with large values for unassignable tasks to simulate infinity. Here, row and column reductions are made within the cost matrix to streamline the optimization process by focusing on iterative refinement through dual-primal optimization such that tasks are strategically reassigned to minimize the total costs until no further reductions are possible and an optimal solution is obtained. This modification improves the algorithm's efficiency, particularly for linear sum assignment problems.

## 2.5. Geostatistical anisotropy

Anisotropy is a prevalent term in different scientific domains with similar connotations; in the earth sciences, in particular, the field of geostatistics, anisotropy describes the directional dependence of a variable's spatial continuity, examples of such variables include, rock and facies types, porosity, permeability, gold grades, etc. Anisotropy is a measure that defines how properties differ with respect to the direction of the measurements made. Mathematically, anisotropy, $\alpha$, is defined by an angle of orientation indicative of the major direction of spatial continuity, and its major and minor range parameters. Matheron (1962, 1963) introduces the term geometric anisotropy and conceptualizes anisotropy through the analogy of an ellipse (in 2D), where the primary direction of anisotropy aligns with the ellipse's semi-major axis, and the secondary, orthogonal direction of continuity aligns with the semi-minor axis. This analogy is particularly apt in geology, where natural processes often extend rock and petrophysical attributes along preferred orientations, rendering these attributes anisotropic due to their varying spatial continuity across different directions. Traditionally, to model anisotropy in a geological setting two definitions are required: the direction of spatial continuity and the anisotropic ratio. Based on the scope of our work, we focus on the latter, anisotropic ratio, a measure that captures the magnitude of anisotropy (Eq. (2)). This quantification provides a scalar value encapsulating the directional dependence of variability within a dataset.

$$\beta = \frac{\alpha_{maj}}{\alpha_{min}} \tag{2}$$

where $\beta$ is the anisotropic ratio, $\alpha_{maj}$ is the relative range of anisotropy in the major direction, $\alpha_{min}$ is the relative range of anisotropy in the minor direction, which corresponds to the width and height of the ellipse respectively. $\beta > 1$ indicates pronounced elongation along the major axis of the ellipse and pronounced elongation along the minor axis when $\beta < 1$. Both scenarios show the existence of anisotropy with respect to the property or datasets considered. Meanwhile, when $\beta = 1$ there is no anisotropy (i.e., isotropy), which indicates the property or datasets measured does not change with direction.

## 3. Methodology

Our proposed generalizable workflow to evaluate stability applies to latent feature spaces from all deep learning approaches for dimensionality reduction. This workflow applies to both image and tabular datasets and comprises the following steps:

1. Preprocess and standardize the predictor features in high-dimensional space to obtain the input data, $X$.
2. Train a deep learning method of choice with a dimensionality reduction architecture and loss function for a large number of model realizations, $k$.
3. Check if the response feature, $Y$, is categorical or continuous. If categorical with different classes, $g$, continue to step (4), otherwise go to step (6).
4. Perform k-means clustering and predict the classes in the response feature, $Y_g^{(pred)}$, while enforcing consistent class label assignments via the modified Jonker-Volgenant algorithm in each latent feature space, $Z^{(k)}$, $\forall\ k$.
5. Compute the percentage change in class label assignment, $\eta$, with respect to truth labels, $Y_g^{(true)}$, for every sample point in $Z^{(k)}\ \forall\ k$.
6. Determine the convex hull polygon to identify the anchor set data indices, $\{I_{A_n}\}$, in every latent feature space, $Z^{(k)}$, and calculate the Jaccard dissimilarity, $\Omega_{\text{jaccard}}$, between each $\{I_{A_n}\}^{(k)}$.
7. In $Z^{(k)}\ \forall\ k$, compute the sequential percentage change in anchor points, $\varepsilon$, between consecutive model realizations.
8. Compute the proposed adjusted stress metric, $\sigma_{adj}$, as a distortion measure between each $Z_k$.
9. Calculate the proposed anisotropy ratio-based measures, $\beta$, in $Z^{(k)}\ \forall\ k$ and associated sequential percentage change in anisotropies, $\delta$, as a measure of data dispersion.
10. Quantitatively compare and visually analyze all stability metrics, $\eta$, $\varepsilon$, $\delta$, and $\sigma_{adj}$, $\Omega_{\text{Jaccard}}$, to assess latent feature space stability.

The first step in our proposed workflow to evaluate the stability of deep learning latent feature spaces is to preprocess and standardize or normalize the predictor features of interest, $P$, $\forall\ j = 1,\dots,J$, where $J = \{1, 2, \dots, \mathbb{N}\}$, and $P \in \mathbb{R}^{N \times d}$. The $P$ predictor features are standardized or normalized in agreement with an appropriate activation function, forming the curated input data, $X$, for a sample size of $N$. A deep learning model is trained with an appropriate architecture and loss function across a sufficient number of epochs to minimize errors and mitigate overfitting when deriving the latent feature space, $Z$. To prevent bias during batch-based model training, the input data, $X$, is shuffled and the corresponding indices are tracked across $Z^{(k)}$ to ensure uniform comparisons across different model realizations. Next, a large number of trained deep learning model realizations, $K$, are generated by varying the random state parameter, which alters the model's weight initializations while maintaining a consistent architecture. This process yields an ensemble of latent feature spaces, $Z^{(k)} \in \mathbb{R}^2$ for each $k = 1, \dots, K$, where $K$ represents a specific subset of natural numbers for model realizations considered. After training, the latent feature space in $Z^{(k)}$ for each $k$ is reordered according to the original indices of $X$, denoted as $I_X$, to ensure that despite the shuffling during training, the deep learning latent feature spaces remain coherently mapped to the initial data samples in $X$.

Pyrcz and Deutsch (2001) introduced a visual metric to identify artifacts near conditioning data, which is pivotal when gauging the effects of data conditioning on probability field simulations for geostatistical realizations. This metric evaluates the data indices' distribution across multiple realizations; a uniform distribution of indices signifies no artifact, whereas unimodal or multimodal Gaussian distributions suggest their presence. Extending this framework, we select a data sample with a consistent index, $i$, in the ensemble of latent feature spaces $\mathbf{Z}^{(k)} \in \mathbb{R}^2$ for every model realization. Then, track the variations in the data sample's position along the x-axis of the normalized latent space, $z_1^{\{i,k\}}$. An approximate uniform distribution of $z_1^{\{i,k\}}$ implies complete sample instability in the data sample, a unimodal or multimodal distribution indicates partial sample stability, while a Dirac-delta function indicates complete sample stability for each realization, where $k = 1, \ldots, K$, and $K = \{1,2,\ldots,\mathbb{N}\}$ represents the specific set of natural numbers for model realizations considered.

For categorical $\mathbf{Y}$ comprising of various classes denoted as $\mathbf{Y}_g$, k-means clustering (Macqueen 1967) is applied in $\mathbf{Z}^{(k)}$, $\forall\, k = 1, \ldots, K$. This step predicts the classes in $\mathbf{Y}$, denoted as $\mathbf{Y}_g^{pred}$, in each $\mathbf{Z}^{(k)}$, yet the absolute choice of category labels with k-means clustering is arbitrary, and are maximally aligned between the predicted $\mathbf{Y}_g^{pred}$ and true class labels, $\mathbf{Y}_g^{true}$, using the modified Jonker-Volgenant algorithm for linear sum assignment with no initialization (Crouse 2016) to ensure consistent class representation across all realizations. To quantitatively assess stability within $\mathbf{Z}^{(k)}$, the percentage change in class label assignment, $\boldsymbol{\eta}$, is calculated for each sample in Eq. (3). This measure tracks variations in $\mathbf{Y}_g^{pred}$ relative to $\mathbf{Y}_g^{true}$, providing an indicator of inferential stability in the latent feature space across all $k$ model realizations considered.

$$\boldsymbol{\eta} = \sum_{i=1}^{N} \mathbb{I}\left(\mathbf{Y}_{g,i}^{true} \neq \mathbf{Y}_{g,i}^{pred}\right) \times 100 \tag{3}$$

where $\mathbf{Y}_{g,i}^{true}$ is the true class label for the $i^{th}$ sample, and $\mathbf{Y}_{g,i}^{pred}$ is the predicted class label for the $i^{th}$ sample, obtained after k-means clustering and alignment using the modified Jonker-Volgenant algorithm. $N$ is the total number of samples in the dataset i.e., sample size, and $\mathbb{I}$ is the indicator function which is 1 if $\mathbf{Y}_{g,i}^{true} \neq \mathbf{Y}_{g,i}^{pred}$, and 0 otherwise.

The quickhull algorithm (Barber et al. 1996) is employed within a latent feature space, $\mathbf{Z}$, to find the convex hull, identify the set of anchor points, $\{\mathbf{A}_n\}$, and its corresponding set of indices represented by $\{I_{\mathbf{A}_n}\}$, where $n$ represents the total number of points defining the hull. For an ensemble of latent feature spaces, $\mathbf{Z}^{(k)}$, both the anchor points, $\{\mathbf{A}_n\}^{(k)}$, and their indices, $\{I_{\mathbf{A}_n}\}^{(k)}$, are determined $\forall\, k$. To evaluate the similarity between the latent feature spaces of $K$ model realizations, we construct Jaccard's dissimilarity matrix, $\boldsymbol{\Omega}_{\text{Jaccard}}$, in a combinatorial manner by systematically comparing the sets of anchor point indices (Jaccard 1912). The Jaccard dissimilarity between an $i^{th}$ and $j^{th}$ model with latent feature spaces $\mathbf{Z}^{(i)}$ and $\mathbf{Z}^{(j)}$ for $i, j \in K$, and $K = \{1,2,\ldots,\mathbb{N}\}$ with $i < j$, and corresponding anchor set indices $\{I_{\mathbf{A}_n}\}^{(i)}$ and $\{I_{\mathbf{A}_n}\}^{(j)}$, is given by Eq. (4).

$$\Omega_{\text{Jaccard}}^{i,j}\left(\{I_{\mathbf{A}_n}\}^{(i)}, \{I_{\mathbf{A}_n}\}^{(j)}\right) = 1 - \frac{\mathcal{C}\left(\{I_{\mathbf{A}_n}\}^{(i)} \cap \{I_{\mathbf{A}_n}\}^{(j)}\right)}{\mathcal{C}\left(\{I_{\mathbf{A}_n}\}^{(i)} \cup \{I_{\mathbf{A}_n}\}^{(j)}\right)} \tag{4}$$

where $n$ is the total number of points defining each convex hull, and the numerator and denominator represent the cardinalities of the intersection and union of the anchor set indices compared respectively. ∩ and ∪ are operators representing set intersection and union, respectively. Given that Jaccard dissimilarity is based on set operations, it is inherently invariant to the sequence and Euclidean transformations of the items compared. Typically, $\Omega_{Jaccard}^{i,j} = 0$ indicates identical sets, and a value of 1 indicates completely dissimilar sets i.e., completely different sets. However, due to the specific application of Jaccard's dissimilarity on anchor set indices in $\mathbf{Z}^{(k)}$, $\Omega_{Jaccard}^{i,j} = 0$ signifies perfect inferential stability, indicating identical anchor set indices implying consistent model predictions and inferences $\forall\, k$. Conversely, $\Omega_{Jaccard}^{i,j} = 1$ reflects complete inferential instability, indicating entirely dissimilar anchor set indices and suggesting significant variability in model inferences or predictions within the ensemble latent feature spaces.

To further assess the stability of anchor sets in $\mathbf{Z}^{(k)}\ \forall\, k$, the absolute percentage change, $\boldsymbol{\varepsilon}$, between consecutive sets of anchors set indices, $\{\mathbf{I}_{\mathbf{A}_n}\}^{(k)}$, is computed via Eq. (5).

$$\boldsymbol{\varepsilon} = \frac{|\mathbf{I}_{diff}|}{|\mathbf{I}_{union}|} \times 100 \tag{5}$$

$$\mathbf{I}_{diff} = \{\mathbf{I}_{\mathbf{A}_n}\}^{(k)}\ \Delta\ \{\mathbf{I}_{\mathbf{A}_n}\}^{(k-1)}$$
$$\mathbf{I}_{union} = \{\mathbf{I}_{\mathbf{A}_n}\}^{(k)}\ \cup\ \{\mathbf{I}_{\mathbf{A}_n}\}^{(k-1)}$$

where $n$ is the total number of points defining each convex hull. Here, $\{\mathbf{I}_{\mathbf{A}_n}\}^{(k)}$ and $\{\mathbf{I}_{\mathbf{A}_n}\}^{(k-1)}$ represent the indices of the current and previous sets of anchor points for the model realizations, respectively, for $k = 2, \ldots, K$, and $K = \{1, 2, \ldots, \mathbb{N}\}$, where $K$ represents a specific subset of natural numbers for model realizations considered. where $\boldsymbol{\varepsilon}$ is a vector where each entry, $\varepsilon^{(k)}$, represents the absolute percentage change in anchor set indices between the $(k-1)^{th}$ and $k^{th}$ model realization. The operators $\Delta$ and ∪ calculate the symmetric difference and union, respectively, between the indices of consecutive anchor sets for each model realization, $k$.

Inspired by Kruskal's (1964) method of assessing and quantifying dissimilarities between high-dimensional and MDS latent feature spaces via normalized stress, we propose the adjusted stress metric, $\boldsymbol{\sigma}_{adj}$, as a distortion measure. This metric quantifies the dissimilarity and evaluates the structural similarity between pairs of latent feature spaces derived from distinct model realizations, applied combinatorically across all $\mathbf{Z}^{(k)} \in \mathbb{R}^2$, for $k = 1, \ldots, K$, model realizations. Given the latent feature space for all model realizations, $\mathbf{Z}^{(k)}$ is of $\mathbb{R}^2$ dimensionality, Euclidean distances serve as the basis for computing distance metrics. Thus, the adjusted stress between the $i^{th}$ and $j^{th}$ model, with respective latent feature spaces $\mathbf{Z}^{(i)}$ and $\mathbf{Z}^{(j)}$, for $i,\ j\ \in K,\ and\ K = \{1, 2, \ldots, \mathbb{N}\}\ with\ i < j$, is formalized in Eq. (6).

$$\sigma_{adj}^{(i,j)} = \sqrt{\frac{\sum_{p=1}^{N-1}\sum_{q=p+1}^{N}\left(d_{p,q}^{(i)} - d_{p,q}^{(j)}\right)^2}{\sum_{p=1}^{N-1}\sum_{q=p+1}^{N}\left(d_{p,q}^{(i)}\right)\left(d_{p,q}^{(j)}\right)}} \qquad (6)$$

where $\sigma_{adj}^{(i,j)}$ is the adjusted stress between the $i^{th}$ and $j^{th}$ latent feature spaces, $(p,q)$ are unique data pairs within each $\mathbf{Z}^{(k)}$. The terms $d_{p,q}^{(i)}$ and $d_{p,q}^{(j)}$ correspond to the Euclidean distances between points $p$ and $q$ within the latent feature spaces $\mathbf{Z}^{(i)}$ and $\mathbf{Z}^{(j)}$, respectively. Hence, $\sigma_{adj}^{(i,j)} = 0$ signifies perfect structural stability, indicating no distortion between the latent feature spaces and suggests that sample locations within $\mathbf{Z}^{(k)}$ are consistent $\forall\, k$. Conversely, $\sigma_{adj}^{(i,j)} \geq 1$ represents complete structural instability, indicating maximal distortion and significant variability in sample locations within $\mathbf{Z}^{(k)}\;\forall\, k$.

Leveraging Boisvert's (2010) local calculation of geometric anisotropy, based on the geometry of an ellipse, we adopt the use of anisotropic ratio, $\boldsymbol{\beta}$, as a quantitative metric that serves both as a geostatistical and geometry-based statistic. The computation of $\boldsymbol{\beta}$ encompasses a variety of methods, including the minimum volume enclosing ellipse (MVEE), alongside local and global techniques that utilize eigenvector and eigenvalue analyses to capture the magnitude of anisotropy at different scales. These methodologies are designed to quantify the spatial spread, variation and orientation, and modality within latent feature spaces, $\mathbf{Z}^{(k)}$, contributing to the integral metrics for stability evaluation. The Khachiyan Algorithm (Khachiyan 1996) is employed to compute the MVEE within the latent feature spaces, $\mathbf{Z}^{(k)}$, for all model realizations, $k$. However, Khachiyan's algorithm can be computationally expensive for large datasets due to its reliance on matrix operations and linear algebra. To address this, a parallelized adaptation is deployed for large datasets (Snyder 2023) while smaller datasets are processed efficiently on a CPU. Following the computation of the MVEE, the magnitude of anisotropy for all $\mathbf{Z}^{(k)}$ is quantified using Eq. (2) from the background section, denoted as $\boldsymbol{\beta}_{MVEE}$.

Subsequently, the global anisotropic ratio, $\beta_{global}$, within a latent feature space, $\mathbf{Z} \in \mathbb{R}^2$, is calculated in a two-step process, ellipse parameterization and anisotropy magnitude quantification. First, $\mathbf{Z}$, is centered to obtain, $\mathbf{Z}_m$, which then applied to Gaussian kernel density estimation to approximate the probability density function (PDF). To isolate the principal axes of data variability, the covariance matrix of $\mathbf{Z}_m$, $\boldsymbol{\Sigma}$, is computed, followed by its eigenvalue decomposition $\boldsymbol{\Sigma} = \boldsymbol{\psi}\boldsymbol{\Lambda}\boldsymbol{\psi}^T$, where $\boldsymbol{\psi}$ and $\boldsymbol{\Lambda}$ represent matrices of eigenvectors and eigenvalues, respectively. The standard deviations along the principal axes, indicating the ellipse's dimensions are estimated from the sorted eigenvalues, thus determining the ellipse's width and height. The orientation of the ellipse is derived from the principal eigenvector $\boldsymbol{v}^{(1)}$, and its angle of orientation, $\theta$, is calculated using $\theta = \arctan\left(v_{z_2}^{(1)}, v_{z_1}^{(1)}\right)$, where $v_{z_1}^{(1)}$ and $v_{z_2}^{(1)}$ are the components of the principal eigenvector along the x and y axes of latent space, $\mathbf{Z}$, respectively. Global anisotropic ratio is determined using the ellipse's width and height, as specified by Eq. (2) in the background section, for all ensemble latent feature spaces $\mathbf{Z}^{(k)} \in \mathbb{R}^2$, where $k = 1, \ldots, K$ enumerates the model realizations. An example calculation for the global anisotropic ratio in the latent feature space is shown for a particular model realization in Figure 1.

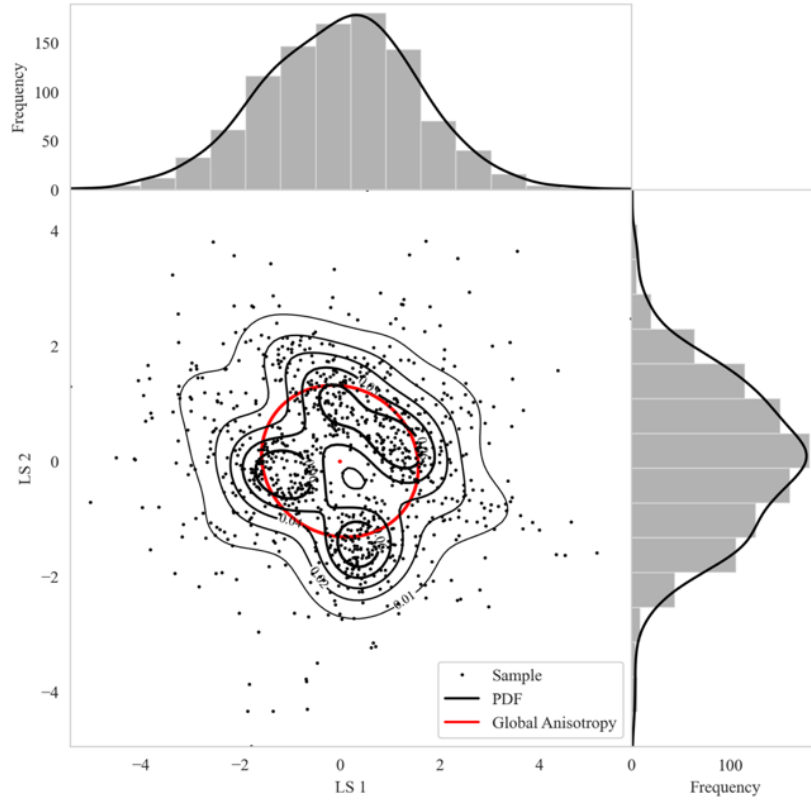

**Fig. 1** Global anisotropic ratio computation from latent feature space for an example model realization with $\beta_{global} = 1.22$

A potential limitation of this method arises when $\mathbf{Z}^{(k)}$ exhibits a multimodal PDF, suggesting clustering or the presence of distinct groups within $\mathbf{Z}^{(k)}$, which could result in a misinterpretation of global anisotropy. This issue, while not identical to Simpson's paradox, a statistical phenomenon where a statistical measure between two features in a population emerges, disappears, or reverses when the population is divided into subpopulations, can lead to misleading aggregated statistics (Yule 1903; Simpson 1951). Consequently, we introduce the local anisotropic ratio as a statistical measure to address this concern.

Similar to the global anisotropic ratio, the local anisotropic ratio, $\beta_{local}$, is derived from the eigen-decomposition of the centered covariance matrix to identify the principal axes of data variability. This decomposition yields the eigenvectors and eigenvalues required to calculate the orientation, width, and height of the ellipse, which are then utilized to calculate the magnitude of anisotropy via Eq. (2) in the background section. This calculation is performed for each latent feature space, $\mathbf{Z}^{(k)} \in \mathbb{R}^2$, where $k = 1, \ldots, K$, and $K = \{1,2,\ldots,\mathbb{N}\}$ represents the set of model realizations. The principal distinction between global and local anisotropic ratio computation lies in the determination of the PDF, for local anisotropic ratio, a Gaussian kernel density estimate is applied to $\mathbf{Z}_m$, and contour levels at a 95% confidence interval are established, delineating regions of significant data density. Within each contiguous region, the methodology for global anisotropic ratio determination is applied, resulting in localized ellipse dimensions (i.e., width and height) shown in Figure 2 using the same example realization from previous global anisotropic ratio calculation.

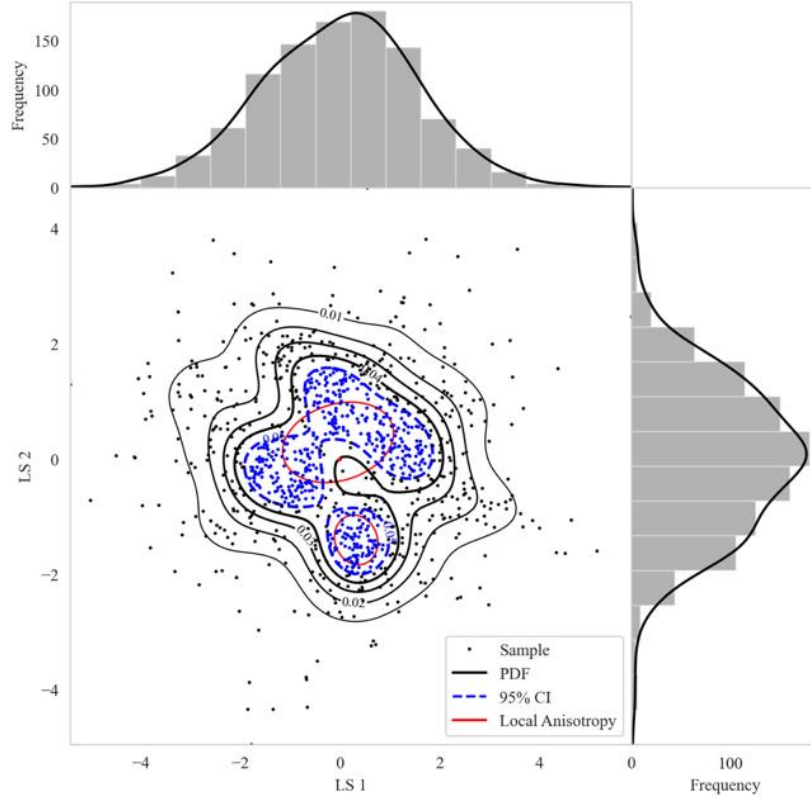

**Fig. 2** Local anisotropic ratio computation from latent feature space for the same example model realization with two regions of significant data density having $\beta_{local}^{(1)} = 1.70$ and $\beta_{local}^{(2)} = 1.12$

However, given the potential for multiple sets of ellipse dimensions corresponding to the number of identified regions, a unified measure is necessary to encapsulate the variations in local anisotropy across each $\mathbf{Z}^{(k)}$ latent feature space. Consequently, we propose the harmonic mean of local anisotropic ratio values, calculated using Eq. (7) for an ensemble of latent feature spaces, $\mathbf{Z}^{(k)}$. The harmonic mean is employed to minimize the influence of extreme anisotropic ratios, acknowledge spatial variation, and ensure the dependency of anisotropy within each region is captured, thereby providing a more representative measure of local anisotropy.

$$\boldsymbol{\beta}_{harmonic} = \frac{n_{local}}{\sum_{j=1}^{n_{local}} \left( \frac{1}{\beta_{local}^{(j)}} \right)} \quad (7)$$

where $n_{local}$ is the number of delineating regions with significant data density within local anisotropic ratio computations for the latent space. where $j$ is the index of summation that runs from 1 to $n_{local}$, and $\beta_{local}^{(j)}$ is the local anisotropic ratio value within each region, and $\boldsymbol{\beta}_{harmonic}$ is a vector representing the harmonic mean of all local anisotropic ratio values across the latent feature spaces, $\mathbf{Z}^{(k)}$.

To summarize the proposed anisotropy-based measures as indicators of stability, we employ the use of Eq. (8) to calculate the absolute percentage change, $\delta$, between consecutive model realizations for all anisotropy-based measures.

$$\delta = \frac{|\beta^{(k)} - \beta^{(k-1)}|}{|\beta^{(k-1)}|} \times 100 \tag{8}$$

where $\beta^{(k)}$ and $\beta^{(k-1)}$ represents the magnitude of anisotropy for the current and previous model realizations, respectively, which may pertain to MVEE, global, or local anisotropic ratios. The calculation is performed for each model realization, where $k = 2, \ldots, K$, and $K = \{1, 2, \ldots, \mathbb{N}\}$ where, $K$ represents a specific subset of natural numbers for model realizations considered. $\delta$ is a vector where each entry, $\delta^{(k)}$, represents the absolute percentage change in anisotropy-based stability measures between the $(k-1)^{th}$ and $k^{th}$ model realization.

Based on the proposed set of stability metrics, there are three types of stability to be evaluated within latent feature spaces, i.e., sample, inferential, and structural stability. A concise overview of these metrics, including their mathematical definitions and interpretations is included in Table 1. First, we define the terms used in this table to facilitate a brief understanding of our methodology and findings:

1. Total instability: This indicates a complete lack of predictability in sample locations within the latent feature space, evidenced by a uniform distribution in histogram visuals.

2. Partial instability: This suggests a moderate level of predictability in sample location with some randomness, shown by unimodal or multimodal distributions in the histogram visuals.

3. Total stability: This denotes absolute predictability with no variation in sample locations, characterized by a Dirac-delta function.

The adjusted stress, $\sigma_{adj}$, and Jaccard dissimilarity, $\Omega_{\text{Jaccard}}$, further quantifies stability. Lower values of $\sigma_{adj}$ and $\Omega_{\text{Jaccard}}$ indicate higher stability, with ranges defined as follows:

1. Significant stability: An extremely low level of distortion or dissimilarity ($\sigma_{adj}, \Omega_{\text{Jaccard}} < 0.2$), indicating very consistent structures or inferences in the latent feature space.

2. Partial stability: A low to moderate level of distortion or dissimilarity ($0.2 \leq \sigma_{adj}, \Omega_{\text{Jaccard}} < 0.5$), suggesting generally consistent structures or inferences with some level of variability in the latent feature space.

3. Instability: A moderate to high level of distortion or dissimilarity ($0.5 \leq \sigma_{adj}, \Omega_{\text{Jaccard}} < 0.7$), indicating noticeable inconsistences in structures or inferences in the latent feature space.

4. Significant instability: A high level of distortion or dissimilarity ($\sigma_{adj}, \Omega_{\text{Jaccard}} \geq 0.7$), reflecting major inconsistencies and variability in structures or inferences between latent spaces.

**Table 1** Stability metrics summary

| Stability Type | Stability Metric | Indicators | Interpretation |
|---|---|---|---|
| Sample | Histogram visuals | Uniform distribution | Total instability |
| | | Unimodal or multimodal distribution | Partial instability |
| | | Dirac-delta function | Total stability |
| Structural | Adjusted stress, $\sigma_{adj}$ | < 0.2 | Significant stability |
| | | 0.2 to < 0.5 | Partial stability |
| | | 0.5 to < 0.7 | Instability |
| | | $\geq 0.7$ | Significant instability |
| Inferential | Jaccard dissimilarity, $\Omega_{\text{Jaccard}}$ | < 0.2 | Significant stability |
| | | 0.2 to < 0.5 | Partial stability |
| | | 0.5 to < 0.7 | Instability |
| | | $\geq 0.7$ | Significant instability |

## 4. Results and Discussions

The proposed workflow to evaluate the stability of latent feature spaces in deep learning employs an autoencoder trained across two distinct scenarios, synthetically generated datasets exhibiting varying levels of multivariate correlation, and a real-world dataset to verify the workflow's practicality.

The autoencoder model comprises an input layer corresponding to the data's dimensionality, followed by an encoder with two sequential linear layers, each with 128 neurons and mean squared error (MSE) as the model's loss function. The encoder compresses the input layer to a two-dimensional latent feature space employing LeakyReLU activation functions. Symmetrically, the decoder reconstructs the input from this latent feature space, also using LeakyReLU activations. The training parameters include a batch size of 16, a learning rate of 0.001, and the Adam optimizer over 10,000 epochs while minimizing the MSE, which is determined experimentally to be sufficient for training convergence. The epoch count used is considered optimal up to the point where a plateau is observed in the MSE, beyond which no significant reduction in model learning error occurs. Stability evaluation is conducted across $K = 500$ distinct autoencoder realizations, each with different weight initializations set by unique random states. The dataset is shuffled for each training instance, with index tracking ensuring consistent latent space mapping post-training. Afterward, the reordering of latent spaces aligns with the dataset's original indices, allowing for coherent comparative analysis across the ensemble of latent feature spaces.

### 4.1. Workflow demonstration with benchmark cases

To investigate the effects of correlation strengths in datasets with respect to latent feature space stability for autoencoders, two synthetic datasets are generated via a modified version of a publicly available synthetic data generation workflow (Pyrcz 2023) as benchmark cases. The first dataset, characterized by negligible inter-variable correlation, represents a low-correlation scenario. The second dataset exhibits moderate to high correlation, with specific positive and negative correlations structured between the first three predictors, and the fourth predictor remaining uncorrelated. Both datasets contain four predictor features sampled from a multi-Gaussian distribution and a categorical response with four classes, each with a sample size of 1000. The intuition behind using these low and moderately high correlated datasets as benchmark cases is to show the effects of stability in the latent feature space when an autoencoder cannot easily compress the data (low correlation) and when it can easily compress it (moderately high to high correlation).

Applying the proposed workflow to the low correlation dataset, an ensemble of distinct latent feature spaces is constructed from $k = 500$ autoencoders using the previously discussed architecture during training. To ascertain training efficacy, six autoencoder realizations were randomly chosen for detailed examination of their latent feature spaces, and the stability of two sample points in all realizations is evaluated. Using the visual metric adapted from Pyrcz and Deutsch (2001), a data sample with index 890 is selected and tracked across the normalized x-axis of the ensemble of latent feature spaces (i.e., LS 1) for all autoencoder realizations, which has an approximate uniform distribution shown in Figure 3, indicating complete sample instability of the data sample considered in these latent spaces. Similarly, data sample index 45 is tracked and found to have a unimodal Gaussian distribution, indicating partial sample instability.

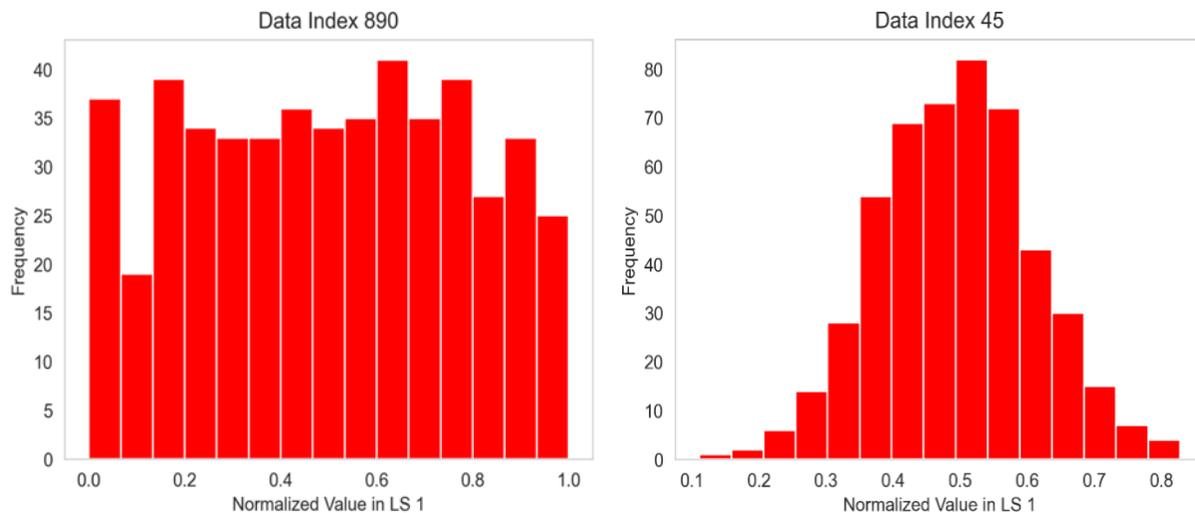

**Fig. 3** Example data index tracked in the x-axis of the normalized latent feature space across autoencoder realizations in the low correlation dataset. Left: Data index 890 suggesting complete sample instability. Right: Data index 45 suggests partial sample stability

The six realizations depicted in Figure 4 illustrate the diversity in the latent spaces as a consequence of initializing the autoencoders with different random states. The convex hull polygon represented by the red dashed lines demarcates the outer bounds of the data distribution and identifies the anchor set within each latent space. The predicted class labels, $\left(Y_g^{pred}\right)^{(k)}$, determined via k-means and refined by the modified Jonker-Volgenant algorithm, are assigned uniform color codes for visual consistency across all autoencoder realizations. To ensure that the autoencoder's model error is not confounded with any instability type, the corresponding MSE training losses are inspected in Figure 5, which shows the MSE plateaus around 9,000 epochs with a minimum loss of 0.107 for all autoencoder realizations.

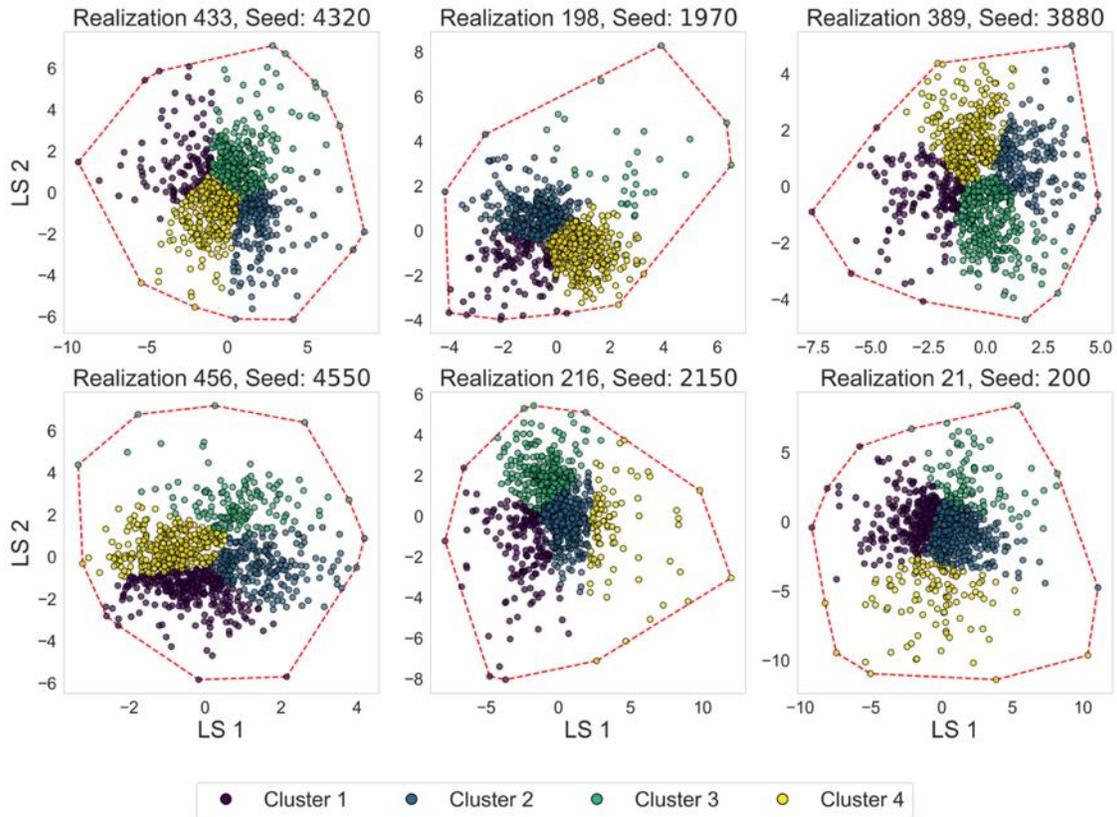

**Fig. 4** Latent feature space for six example autoencoder realizations color-coded by the predicted class label within the categorical response feature in the low correlation dataset. The red dashed line is the convex hull polygon consisting of the anchor set in each realization

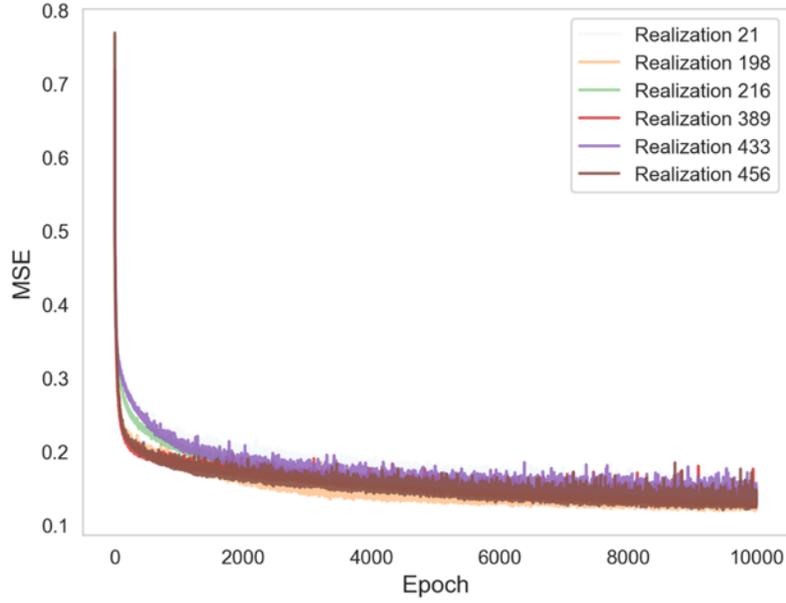

**Fig. 5** MSE training error visualization for six autoencoder model realizations for the low correlation dataset

To assess the stability of the predicted class labels, $\left(Y_g^{pred}\right)^{(k)}$, the percentage change in class label assignment, $\eta$, is monitored for each data sample relative to its truth class assignment across all autoencoder realizations. To evaluate the stability of the anchor sets for all autoencoder realizations, the percentage change, $\varepsilon$, between consecutive sets of anchors set indices, $\{I_{A_n}\}^{(k)}$, is observed in the low correlation dataset. Figure 6 shows these stability metrics as a measure of inferential instability via histograms with the percentile estimates at 10%, 50%, and 90% represented by P10, P50, and P90 respectively as an uncertainty measure.

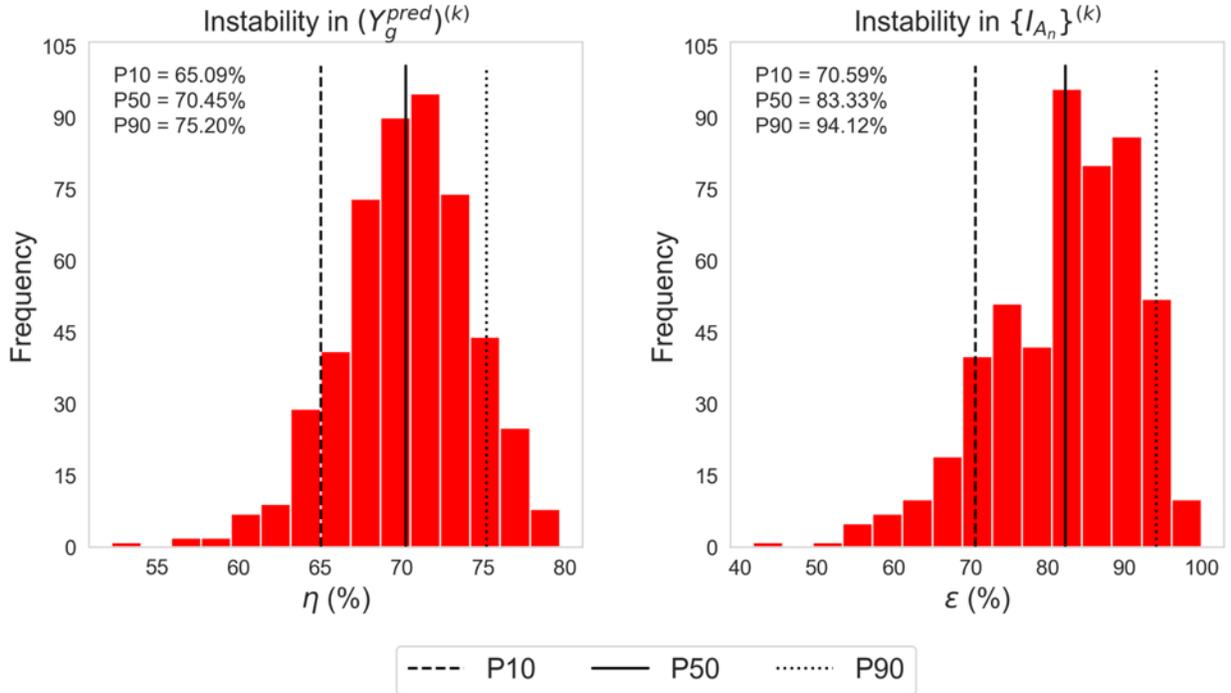

**Fig. 6** Percentage changes as a measure of instability in low correlation dataset for all autoencoder realizations. Left: Predicted class labels within the categorical response feature for each data sample. Right: Consecutive anchor sets indices. where P10, P50, and P90 are the percentile estimates at 10%, 50%, and 90% for each measure

Additionally, our proposed adjusted stress metric and Jaccard dissimilarity are applied as stability measures in a combinatorial manner. Upon estimation, both metrics' matrices are sorted using a hierarchical agglomerative clustering method with Ward linkage to enable the visualization of stability or instability groupings between all autoencoder realizations. Figure 7 shows the sorted lower triangular matrices for adjusted stress, $\boldsymbol{\sigma}_{adj}$, and Jaccard dissimilarity, $\boldsymbol{\Omega}_{\text{Jaccard}}$, excluding the diagonals, along with its respective estimated PDF. At a glance, the sorted adjusted stress metric for most autoencoder realizations have values greater than 0.4 and a mode of ~0.67 indicating structural instability between latent feature spaces. Meanwhile, for most autoencoder realizations, the Jaccard dissimilarity values are greater than 0.6 and a mode of ~0.86 indicating dissimilarity between anchor set indices and significant inferential instability between the ensemble of latent feature spaces.

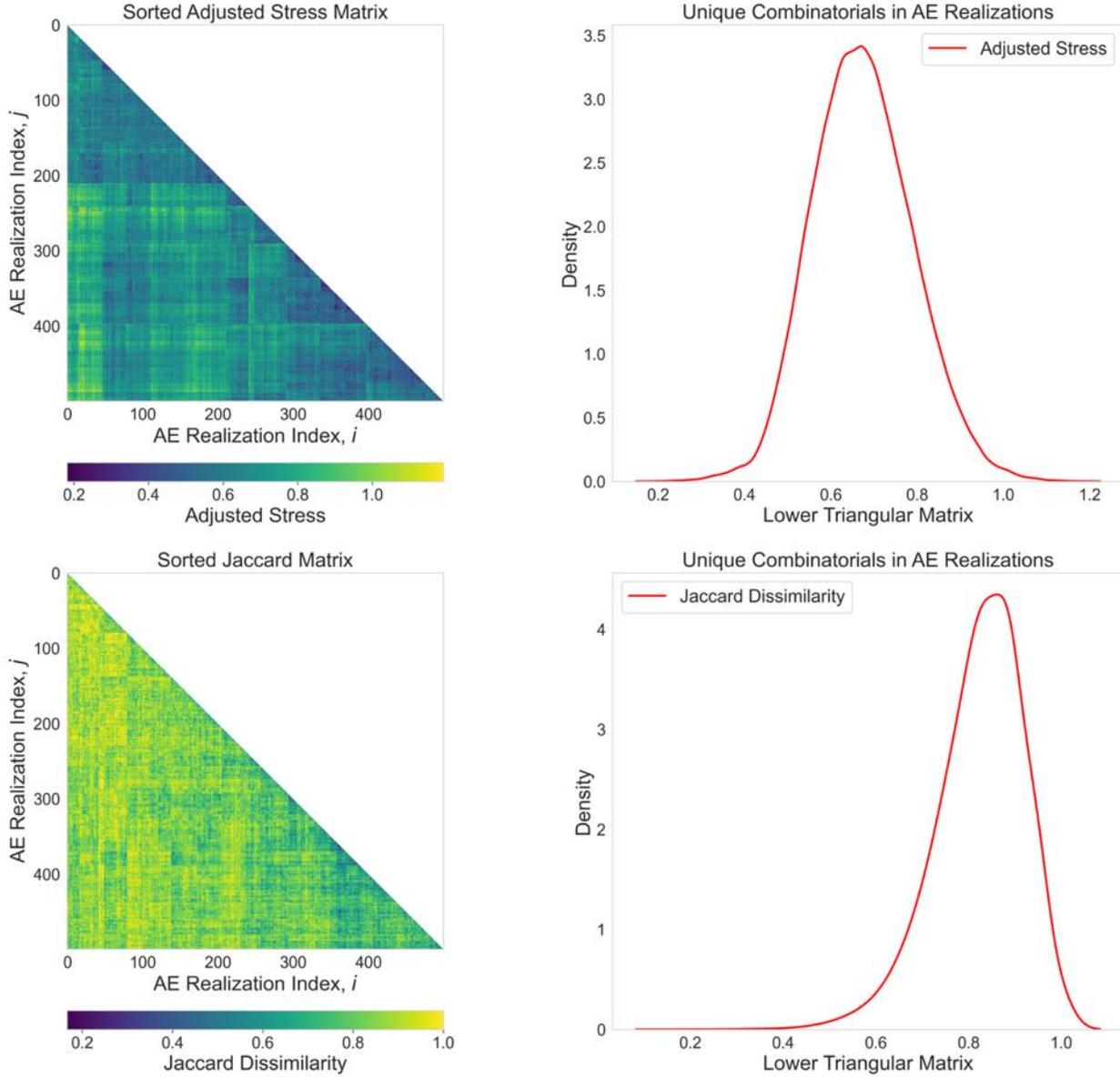

**Fig. 7** Lower triangular matrix excluding the diagonal for all combinations of autoencoder realizations in the low correlation dataset, along with its estimated PDF. Top: Adjusted stress between latent feature spaces. Bottom: Jaccard dissimilarity between anchor set indices

Next, we evaluate stability using our proposed MVEE, global, and locally computed anisotropic ratios (i.e., $\beta_{MVEE}$, $\beta_{global}$, $\beta_{local}$) computed from the ensemble latent feature space for all autoencoder realizations as a measure of spatial variation. The percentage change, $\delta$, between consecutive anisotropic ratios in the latent feature spaces for all autoencoder realizations is shown in Figure 8 with the percentile estimates at 10%, 50%, and 90% represented by P10, P50, and P90 respectively as an uncertainty measure.

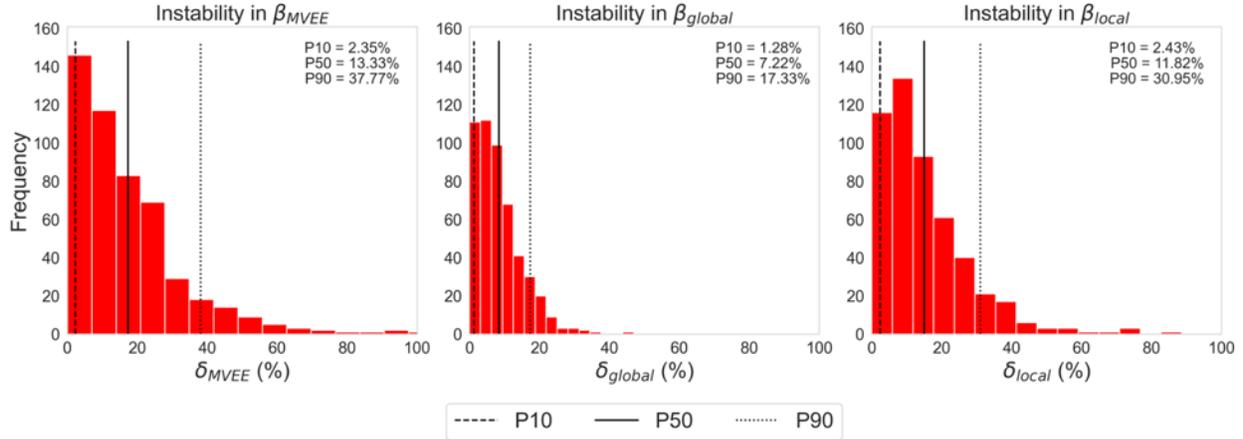

**Fig. 8** Percentage changes in consecutive anisotropic ratio computation in autoencoder latent spaces for all realizations for the low correlation dataset. Left: Minimum Volume Enclosing Ellipse (MVEE). Center: Global method. Right: Local method

The proposed workflow is replicated for the benchmark case with a moderately high correlated dataset. During model training, the MSE reaches a plateau at approximately 8,000 epochs, with a consistent minimum loss of ~0.03 across all autoencoder realizations. The results are summarized in Tables 2 and 3. A comparative analysis of the stability metrics obtained from the moderately high-correlated dataset and the previously discussed low correlation dataset reveals intriguing insights. As the correlation between predictors within the input dataset increases, we observe a corresponding increase in stability of the latent feature space. This suggests that a higher degree of correlation leads to more stable representations within the latent space. However, our findings show regardless of the initial correlation levels among predictor features, a notable degree of instability persists within the latent feature spaces.

**Table 2** Proposed percentage-based stability metrics for all autoencoder realizations in the moderately high correlated dataset.

| Stability Metrics | P10 | P50 | P90 |
|---|---|---|---|
| $\eta$ (%) | 41.80 | 50.95 | 60.71 |
| $\varepsilon$ (%) | 30.69 | 46.15 | 60.00 |
| $\delta_{MVEE}$ (%) | 2.59 | 14.28 | 35.05 |
| $\delta_{global}$ (%) | 1.53 | 8.49 | 18.95 |
| $\delta_{local}$ (%) | 2.88 | 14.91 | 34.89 |

**Table 3** Proposed combinatorial-based stability metrics for all autoencoder realizations in the moderately high correlated dataset.

| Stability Type | Stability Metrics | Mode | Interpretation |
|---|---|---|---|
| Structural | $\sigma_{adj}$ | 0.20 | Significant stability |
| Inferential | $\Omega_{\text{Jaccard}}$ | 0.48 | Partial stability |

*4.2. Case study demonstration*

The demonstrated workflow is repeated on the publicly available wine dataset (Aeberhard and Forina 1991) with 13 predictors and a categorical response feature with 3 classes and a sample size of 178. During training, the MSE reached a plateau at approximately 9,000 epochs, with a consistent minimum loss of ~0.07 across all autoencoder realizations. The results from applying our proposed workflow are summarized in Tables 4 and 5. Table 5 indicates reduced distortion between latent spaces and partial structural stability due to its adjusted stress of 0.27. However, the Jaccard dissimilarity between the anchor sets indices is 0.71, meaning the anchor set indices change significantly, suggesting inferential instability.

**Table 4** Proposed percentage-based stability metrics for all autoencoder realizations in the wine dataset.

| Stability Metrics | P10 | P50 | P90 |
|---|---|---|---|
| $\eta$ (%) | 7.87 | 11.80 | 20.79 |
| $\varepsilon$ (%) | 53.85 | 70.59 | 82.35 |
| $\delta_{MVEE}$ (%) | 2.16 | 12.17 | 27.64 |
| $\delta_{global}$ (%) | 1.24 | 6.15 | 15.50 |
| $\delta_{local}$ (%) | 2.81 | 19.08 | 56.90 |

**Table 5** Proposed combinatorial-based stability metrics for all autoencoder realizations in the wine dataset.

| Stability Type | Stability Metrics | Mode | Interpretation |
|---|---|---|---|
| Structural | $\sigma_{adj}$ | 0.27 | Partial stability |
| Inferential | $\Omega_{\text{Jaccard}}$ | 0.71 | Significant instability |

Comparing the sorted adjusted stress in Figure 9 for the three datasets: low correlation, moderately high correlated, and wine datasets over all autoencoder realizations, observing the modes of 0.67, 0.20, and 0.27, respectively. This implies between latent feature spaces, the low correlation benchmark case (red) is distorted and has structural instability, the moderately high correlation benchmark case (in green) has reduced distortion and significant structural stability, while the wine dataset (blue) has partial structural stability. Figure 10 shows for most autoencoder realizations, the Jaccard dissimilarity between anchor set indices has the modes of 0.86, and 0.48 for the low correlation (red) and moderately high correlated (green) datasets over all autoencoder realizations, respectively. This indicates significant dissimilarity and inferential instability, and partial similarity and stability in the respective datasets. For the wine dataset (blue) the Jaccard dissimilarity is 0.71, implying significant dissimilarity and inferential instability exists.

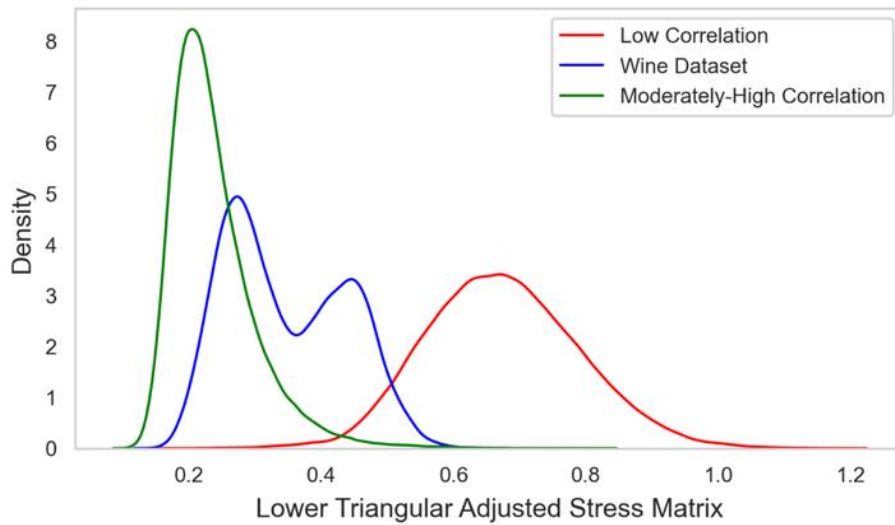

**Fig. 9** Comparison between the lower triangular matrix excluding the diagonal for the adjusted stress metric between latent feature spaces as indicators of stability for all combinations of autoencoder realizations in the low correlation (red), wine (blue), and moderately high correlated datasets (green)

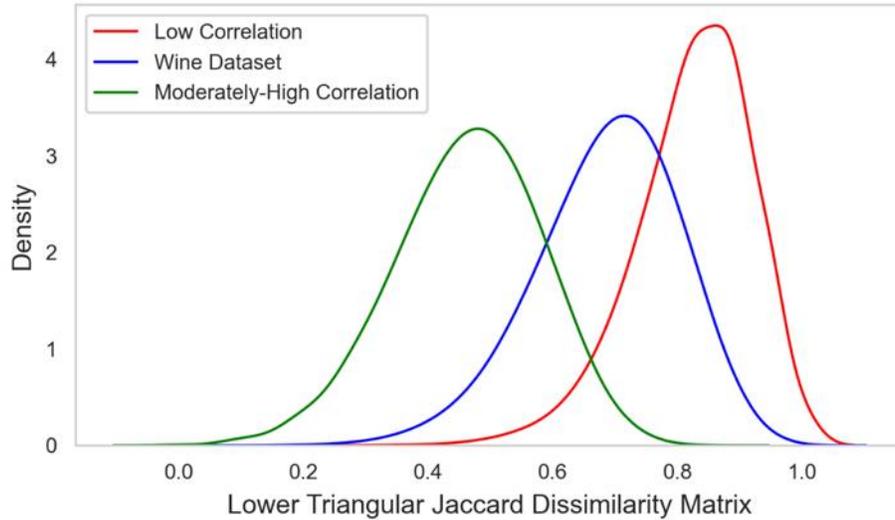

**Fig. 10** Comparison between the lower triangular matrix excluding the diagonal for the Jaccard dissimilarity metric between anchor set indices as indicators of stability for all combinations of autoencoder realizations in the low correlation (red), wine (blue), and moderately high correlated datasets (green)

Lastly, to quantify variation or spread in the set of stability metrics, the percentage changes in class label prediction within the response feature, consecutive anchor set indices, and anisotropic ratio-based measures are compared for the three datasets in Figure 11. Figure 11 shows significant changes in class label predictions, $\eta$, for data samples within the latent feature spaces with the wine dataset having the most stability, even though there is ~13% instability in this measure with some outlying realizations with increased instability, while the low and medium-high correlation datasets have ~75% and 50% instability, respectively. However, comparing the average percentage changes in consecutive MVEE and global anisotropic ratios ($\delta_{MVEE}$ and $\delta_{global}$), in the former, the wine dataset has a reduced spread compared to the others, while the latter has similar variations with a few outlying realizations indicating instability for all datasets. Regardless of the comparable spread in $\delta_{MVEE}$ and $\delta_{global}$, there is an inherent instability of 13.26% and 7.23% on average within all datasets. Comparing the average percentage changes in consecutive local anisotropic ratio, $\delta_{local}$, the wine dataset has the most variation, which may suggest the presence of localized spatial dependencies, while the low and moderately high correlations have similar smaller variations with respect to anisotropy.

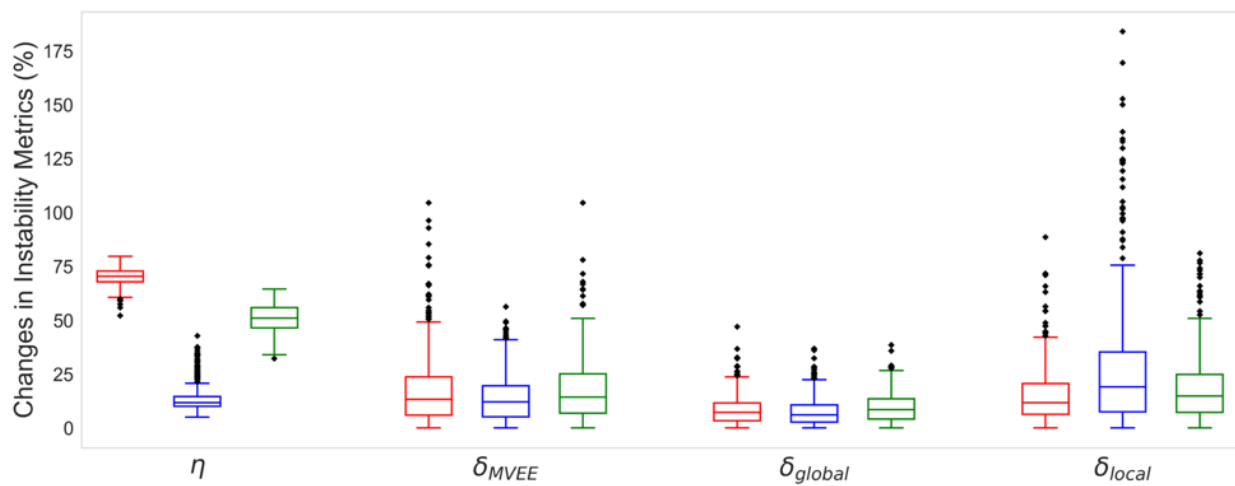

**Fig. 11** Comparison between the percentage change in stability metrics for class label predictions ($\eta$) and anisotropic ratio-based measures ($\delta_{MVEE}$, $\delta_{global}$, and $\delta_{local}$) for all autoencoder realizations in the low correlation (red), wine (blue), and moderately-high correlated datasets (green)

## 5. Conclusions

We introduce a workflow for assessing stability for deep learning latent feature spaces, filling a gap in dimensionality reduction research and applications. Our method is suitable for tabular and image datasets and evaluates sample, structural, and inferential stability, offering a systematic way to quantify and interpret stability in latent feature spaces. Our empirical analysis using an autoencoder and three datasets reveals significant instabilities, challenging the assumption that deep learning inherently ensures stable feature extraction in latent spaces. This has implications for the consistency, reliability, and interpretability of models in areas like bioinformatics, earth sciences, and computer vision.

Our results highlight the need for stability evaluations in latent feature space analysis, setting a new standard for such assessments. Future work should investigate instability mechanisms in deep learning models and develop more robust dimensionality reduction methods. Our proposed workflow has limitations, including potential biases from specific clustering and alignment algorithms. Future research should consider alternative methods and extend stability metric applicability to more deep learning architectures and datasets, refining machine learning methodologies for high-dimensional data analysis.

## Credit author statement

Ademide O. Mabadeje: Data curation, Conceptualization, Methodology, Software, Validation, Visualization, Formal analysis, Writing – Original draft.
Michael Pyrcz: Data curation, Conceptualization, Methodology, Supervision, Funding acquisition, Writing – Reviewing and Editing.

## Declaration of competing interest

The authors declare that they have no known competing financial interests or personal relationships that could have appeared to influence the work reported in this paper.

## Acknowledgments

The authors sincerely appreciate Equinor and the Digital Reservoir Characterization Technology (DIRECT) consortium's industry partners at the Hildebrand Department of Petroleum and Geosystems Engineering, University of Texas at Austin for financial support.

## Data Availability

The data and well-documented workflow used will be publicly available on the corresponding author's GitHub Repository: https://github.com/Mide478/DeepLearning-LatentSpace-StabilityEvaluation on publication.

## ORCID iD

Ademide Mabadeje   https://orcid.org/0000-0002-7594-6517
Michael J. Pyrcz     https://orcid.org/0000-0002-5983-219X